\documentclass[
]{ceurart}

\usepackage{listings}
\begin{document}

\copyrightyear{2026}
\copyrightclause{Copyright for this paper by its authors.
  Use permitted under Creative Commons License Attribution 4.0
  International (CC BY 4.0).}

\conference{EVALITA 2026: 9th Evaluation Campaign of Natural Language
Processing and Speech Tools for Italian, Feb 26 – 27, Bari, IT}

\title{Challenger at MultiPRIDE: Is It Hate Speech or Reclaimed?}


\author[1]{Hadi Bayrami Asl Tekanlou}[%
orcid=0009-0002-7206-6116,
email=h.bayrami1403@ms.tabrizu.ac.ir,
url=https://github.com/HadiBayrami/,
]
\cormark[1]
\address[1]{Department of Computer Science, University of Tabriz,
29 Bahman Boulevard,
Tabriz 51666-16471, Iran
}

\author[2]{Mahdi Bakhtiyarzadeh}[%
orcid=,
email=m.bakhtiyarzadeh1403@ms.tabrizu.ac.ir,
url=https://github.com/Mahdi8424,
]
\address[2]{Department of Computer Science, University of Tabriz,
29 Bahman Boulevard,
Tabriz 51666-16471, Iran
}

\author[1]{Jafar Razmara}[%
orcid=0000-0002-6320-8517,
email=razmara@tabrizu.ac.ir,
url=,
]
\address[]{}

\cortext[1]{Corresponding author.}
\maketitle
\begin{abstract}
The spread of hate speech has become increasingly harmful in modern digital environments, particularly on social networking platforms. While recent advances have shown promising results in automatic hate speech detection, a key challenge remains: distinguishing genuine hate speech from reclaimed language. Accurate labeling is difficult due to the nuanced and context-dependent nature of reclaimed expressions. In this paper, we present a simple and interpretable approach for distinguishing hate speech from reclaimed language, developed for the MultiPride Shared Task. Our method generates dense semantic text embeddings and incorporates a label-noise filtering stage using Cleanlab with logistic regression, followed by a Multi-layer Perceptron (MLP) neural network for final classification. The system is designed to operate under limited computational resources while maintaining strong performance. We evaluate our approach using precision, recall, and F1-score, including macro-averaged metrics. Experimental results demonstrate robust performance despite extreme class imbalance in the dataset. Overall, the findings highlight the potential for further improvements through larger embedding models and more advanced preprocessing techniques while preserving interpretability.
\end{abstract}

\noindent\textbf{Content Warning:} This paper contains examples of explicitly offensive content.



\section{Introduction}

The primary venue for communication between people is now the Internet. This has democratized online discourse, enabling users to share opinions on a vast array of topics. As a result, online platforms have become major spaces for public discourse, enabling rapid information exchange and large-scale interaction among users from diverse backgrounds. While this promotes free speech and broader participation, it also facilitates harmful behaviors when individuals target others based on religion, race, sex, sexual orientation, or political affiliation ~\cite{10848067,schmidt-wiegand-2017-survey,10.1007/s11192-020-03737-6}. The use of such harmful or offensive expressions is referred to as hate speech. The exponential growth of social media platforms has facilitated this spread, transforming how hate speech disseminates through anonymity and rapid sharing ~\cite{schmidt-wiegand-2017-survey,10.1145/3605098.3635964}. Hate speech, as a form of online abuse, has become a focal point in societal discussions, with computational tools emerging to mitigate its impact on individuals and communities ~\cite{Toktarova2023}. On the opposite end of the spectrum, members of the same social group may use terms traditionally considered offensive as part of in-group appropriation or semantic reclamation ~\cite{Zsisku2024IEEE}. This phenomenon, known as reclaimed language, presents a unique challenge in hate speech detection, where terms traditionally used derogatorily are repurposed by marginalized communities, such as the LGBTQ+ group, through mock impoliteness or semantic reclamation ~\cite{Zsisku2024IEEE}. For instance, reclaimed slurs in in-group contexts may not convey hatred but instead foster solidarity; however, automated models often misclassify them as hateful, leading to false positives and restricting self-expression. It can be difficult to differentiate hate speech from other manifestations of potentially harmful language used within a particular group, as meaning depends strongly on speaker identity and contextual usage~\cite{10.1371/journal.pone.0237861}. 

This study makes the following contributions:

\begin{itemize}
    \item We begin by systematically cleaning the raw text data to reduce noise, filtering out URLs, non-ASCII characters (including emojis), and specific user mentions, while carefully stripping the \# symbol from hashtags to retain their semantic textual content.
    \item To ensure reliable evaluation on our highly unbalanced dataset, we split the data into training and validation sets using stratified sampling, preserving the proportional distribution of labels across both subsets.
    \item We implement a comprehensive backtranslation augmentation strategy on the training data using the Google Translate API via deep\_translator. 
    \item We extract deep semantic features from the text samples by generating high-dimensional embeddings using the state-of-the-art intfloat/e5-large-v2 model.
    \item We enhance the quality of our training signal by applying CleanLearning from the cleanlab.classification framework. 
    \item Finally, we design and train a lightweight Neural Network specifically optimized for this task.

\end{itemize}

\section{Related Work}

As presented in the previous section, distinguishing between hate speech and reclaimed language is challenging due to the nature and linguistic complexity, since a clear boundary should be defined ~\cite{Zsisku2024IEEE}. This problem stems from the contextual dependence of language, where marginalized groups, like the LGBTQ+ community, can use terms that are typically used as slurs to promote solidarity through "mock impoliteness" without expressing hatred. These subtleties are frequently overlooked by automated detection systems, which results in false positives that incorrectly label reclaimed language as hateful, restricting self-expression and possibly exacerbating discrimination ~\cite{Zsisku2024IEEE,10.1371/journal.pone.0237861}. Biases in training datasets and classifiers, for example, make this problem worse because models trained on primarily out-group examples may overgeneralize slurs while neglecting in-group reclamation. Linguistic analyses also show how generalized forms of hate speech, which frequently rely on lethal or quantity-laden rhetoric, differ from directed forms, which are aggressive and personal. However, both can overlap with reclaimed uses in online contexts ~\cite{elsherief2018hatelingotargetbasedlinguistic}. Early attempts to identify hate speech relied on rule-based keyword matching or simple machine learning classifiers. While these methods provided some level of detection, they struggled with context, ambiguity, sarcasm and had high error rates for sophisticated language ~\cite{schmidt-wiegand-2017-survey}. Deep learning, with the introduction of semantic embeddings particularly using transformer models (e.g. BERT), has improved performance and made it easier to identify hateful intent across languages and domains by capturing bidirectional context. For example BERT was reweighted to reduce the racial bias in social media by accounting for African-American English (AAE) tweets being overclassified as offensive compared to Standard American English (SAE) tweets ~\cite{10.1371/journal.pone.0237861}. Despite these improvements, the models are still limited in their ability to handle reclaimed language due to their focus on speaker-group dynamics instead of surface level patterns ~\cite{Zsisku2024IEEE}. Recent research shows just how crucial contextual information is for improving detection accuracy. When models take into account conversational threads or topic-specific metadata—for example, from Twitter—they can achieve up to 5.5 points higher Macro F1 scores in multi-label tasks, particularly for pandemic-related hate speech in Spanish varieties ~\cite{10076443}. Context helps clarify meaning: a slur used within an in-group conversation may signal reclamation rather than aggression, and looking at replies or thread structure prevents interpreting posts in isolation. Psycholinguistic work on hate “lingo”—such as name-calling in targeted hate compared to more general religious or group-based rhetoric—also reinforces this need for context. Together, these insights support the development of hybrid models that blend linguistic features with training strategies that are aware of reclamation ~\cite{elsherief2018hatelingotargetbasedlinguistic}. Strategies for bias mitigation are necessary for fair detection and to mitigate against inadvertent suppression of reclaimed language or expression. Techniques such as entropy-based regularization or sample reweighting in fine-tuning have shown efficacy at debiasing classifiers to attenuate the disproportionate flagging of AAE or LGBTQ+ discourse ~\cite{10.1371/journal.pone.0237861,Zsisku2024IEEE}. The establishment of specialized datasets, including the Reclaimed Hate Speech Dataset (RHSD), will be essential to test models in an in-group setting and monitor for safeguards in semantic sensitivity ~\cite{Zsisku2024IEEE}. In addition to creating datasets, surveys of the landscape create a call for discourse within an interdisciplinary framework, to synthesize NLP with sociolinguistics, and enact easier boundaries between hate speech and reclamation, alongside the unraveling of online hate speech ~\cite{schmidt-wiegand-2017-survey}. Still, research gaps remain: most datasets lack a rich compendium of distinct examples of reclamation, and a lack of cross-domain generalization to distinguish bias that can exist across varieties of English ~\cite{10.1371/journal.pone.0237861}. In the future, scholars working on the CMC dimension of hate speech should strive for datasets that are multilingual and intersectional, and that capture linguistic complexity in a global context, extending toward upholding justice to ensure models treat individuals belonging to marginalized groups without excessive over-moderation ~\cite{Zsisku2024IEEE,10076443,schmidt-wiegand-2017-survey}.

\section{Task Description and Dataset}
\subsection{Task}
MultiPRIDE  shared task~\cite{evalita2026overview} aims to identify reclaimed language in LGBTQ+ situations on social media, specifically in tweets that use language tied to LGBTQ+ identity in a possibly harmful way or as language of reclamation. Thus, the task is to identify if it is used as reclaimed, so it is a binary classification task. The shared task is multilingual and includes datasets in Italian, Spanish, and English.  \\
The task provides two tracks:
\begin{itemize}
    \item \textbf{Textual}: Participants are only granted access to the unprocessed text of each tweet. No additional metadata or contextually-relevant signals are provided.
    \item \textbf{Textual + Contextual}: Apart from the text, participants are free to rely upon optional contextual information related to the author, including profile description and any other relevant metadata.
\end{itemize}
Participants are welcome to examine not only models that rely strictly on the text of the tweet, but also models that consider additional contextual information if available (such as metadata regarding the author’s profile). This design facilitates the exploration of the effect contextual signals have on the interpretation of potentially sensitive terms.

\subsection{Data}
For this research, the data set was provided through the Multipride Shared Task: a collection of multilingual Twitter data, that has been annotated specifically for its use of reclaimed language. This data comprises English, Italian and Spanish, has been split into training and development, with each instance representing a tweet and being labelled either reclaimed or not reclaimed language – defining a binary classification scenario. Table~\ref{font-table} outlines how the instances are distributed across the three languages (English, Italian, and Spanish) in terms of their training or developmental status.   Representative examples from the dataset are presented in Table~\ref{rep} .
\begin{table}[h]
\caption{\label{font-table} Statistics of the experimental dataset.}
\begin{center}
\begin{tabular}{|l|l|l|l|}
\hline 
\bf Dataset & \bf English & \bf Italian & \bf Spanish \\ 
\hline
Train & 769  & 814 & 657\\
\hline
Development & 257  & 272 & 219\\

\hline
\end{tabular}
\end{center}
\end{table}

\begin{table}[h]
\caption{\label{rep}Representative examples from the dataset.}
\centering
\begin{tabular}{l p{3.5cm} p{7cm} l}

\hline 
\bf Lang & \bf Biography & \bf Tweet & \bf Reclamation \\ 
\hline
En & N/A  & I use the word tranny all the time…but that’s only in reference to working on my cars. Transgendered, transvestite, and drag queen folk are too fabulous to have their descriptions abbreviated. & Yes\\
\hline
En & N/A  & Actually that’s what a faggot is. Fag is just something that needs to be burnt. & No\\
\hline

It & Certo le circostanze non sono favorevoli& In quanto disabile e frocia questi sono i miei PridePrideMonths. Ma vorrei anche dire che il giorno in cui nel manifesto di un evento lgbtqiatransfemminista verrà citato anche l’antiabilismo oltre ad anti sessismo/obtfobia/razzismo/specismo offro da bere & Yes \\
\hline
It & I veri partigiani furono i primi sovranisti! w la patria! & Ecco, adesso pensate all’iter o in affitto ed al male che fate al bambino branco di finocchi arcobaleno & No \\
\hline
Es & Me llaman feminista, roja y bollera & Buenas tardes a rojos, feministas, republicanos, maricones, bolleras y demás LGTBI \#LGTBI \#pride & Yes \\
\hline 
Es & I live for that energy! & Hace rato pasó una caravana de movimiento LGT…etc y algunos me vieron parado observando y me gritaron que yo también era marica. Órale, bien «respetuosas» estas personas que exigen respeto. \#PrideMonth \#Pride2022 \#LGTBQ & No \\
\hline
\end{tabular}

\end{table}

\section{Proposed Approach}
For the MultiPride hate speech detection task, we constructed a pipeline that does not depend on any specific language (also referred to as language-agnostic). This pipeline includes several components: augmentation of training data, generation of robust embeddings through a neural network classifier, filtering of noisy labels and robust embeddings through a neural network classifier, and a neural network-based classification of hate speech detection, and all components of the pipeline are applied to all datasets (both English, Italian, and Spanish) consistently to mitigate the effect of class imbalance and the presence of noise in social media text.

\subsection{Data Preprocessing and Augmentation}Data cleaning is a technique that utilizes raw text and removes sparsity and non-informative artifacts (non-informative means you cannot determine the origin of the content). For example, we remove URLs, user mentions (but keep the semantic meaning of hashtags), and non-ASCII characters (emoticons, symbols, etc.) using regular expressions. We also apply a back-translation augmentation process to the hate speech datasets to address the significant class imbalance that exists. To do this, we take the hate speech minority class and translate it into multiple pivot languages (such as French, German, Russian, and Persian) and translate it back into the original source language using Google Translate API. This creates synthetic, positive samples of the minority class that enriches the training distribution and allows our models to generalize better among the minority class samples.

\subsection{Feature Extraction and Label Cleaning}Text representations are created by the intfloat/e5-large-v2 Sentence Transformer Model. The intfloat/e5-large-v2 Sentence Transformer is a highly sophisticated model specifically designed to deliver semantic embeddings for input sentences. Input sequences are converted into long vector representations (a 1024-dimensional dense representation) that illustrate the minute bits of context found in the original messages that would not be picked up by a traditional lexical approach. Once the input sequences have been embedded, the initial training data is processed with Confident Learning (cleanlab) to ensure that the quality of the labels in the training data is valid. After the embedding process is completed, Logistic Regression classifiers fit the embeddings into clusters based on the classifier identification of likely errors in the labels (label noise). Any instances classified as exhibiting a high incidence of label problems are removed from the training dataset in order to avoid introducing unwanted label noise during the subsequent final training phase.

\subsection{Model Architecture and Optimization}
\begin{figure*}[h]
    \centering
    \includegraphics[width=1\textwidth]{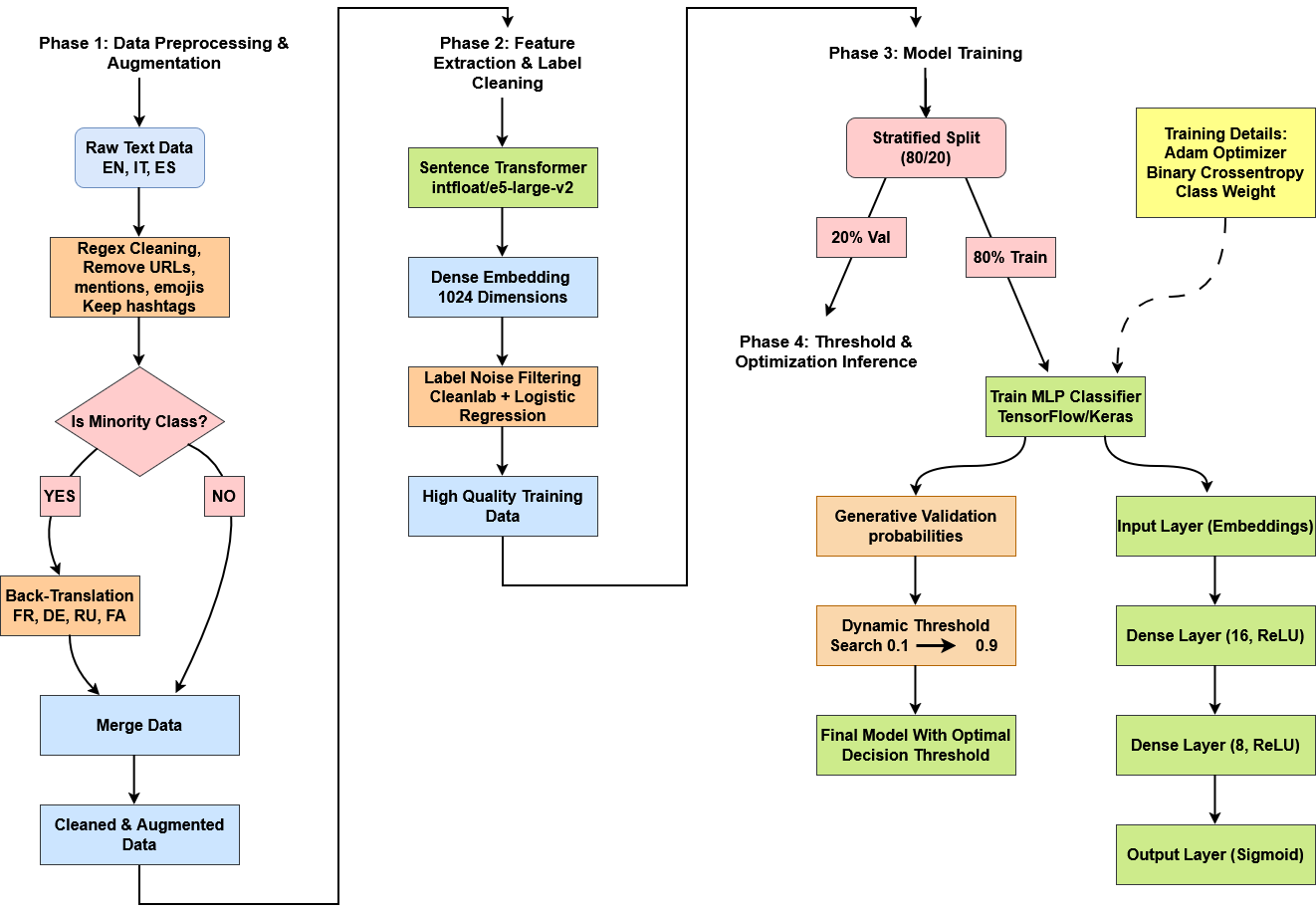}
    \caption{Overview of the proposed system architecture and training pipeline.}
    \label{fig:architecture}
\end{figure*}

Lightweight pipelines utilizing frozen pre-trained language model (PTLM) embeddings along with shallow neural sequence models have been investigated as part of previous studies related to hate speech detection~\cite{3105ac70a55f4725a79468fe8f2c808f}. In this article, authors analyze numerous minimalist architectures incorporated into a "shared task" (combining the use of orderless data) that's built on the combination of PTLMs and LSTMs, BiLSTMs, or transformers with a linear classifier. However, the framework in which we present our findings utilizes one cohesive pipeline created specifically for reclaiming language issues. The Multi-layer Perceptron (MLP) is made using the Tensorflow/Keras framework, and is referred to as the core classifier. The text embedding model we used is \textit{intfloat/e5-large-v2}~\cite{wang2022text}. A multi-layer perceptron consists of an input layer that accepts the sentence embeddings, as well as two dense hidden layers (containing 16 and 8 units, respectively), both of which use the Rectified Linear Unit (ReLU) activation function. Finally, the output layer of the MLP uses the Sigmoid activation function to provide a binary probability score.  The model is trained using the Adam optimizer and Binary Crossentropy loss function. Additionally, to compensate for class imbalance during training, class weights are computed and used to penalize misclassifications of the positive class more severely. The system architecture is illustrated in Figure~\ref{fig:architecture}

\subsection{Validation and Threshold Tuning}In order to maintain representative class distributions throughout the development and evaluation phases, we utilized a 80/20 strata split when creating our training/validation sets. Additionally, after our model is trained, we used an adjusted threshold optimization method instead of simply applying a decision boundary of 0.5. In this method, our system has been designed to search through an entire range of decision boundaries (from 0.1 up to 0.9), and will select the threshold which produced the highest F1 score during validation in order to provide inferred outcomes from our test set using a threshold that was optimized for this project. Thus, we were able to optimize our model for the competition's primary metric while maintaining the tradeoff between precision and recall.

\section{Results}

In this section, we describe the results obtained for multipride 2026. A total of 19 teams participated in
the competition. The results presented in this paper are based on Precision, Recall, F1-score, and along
with their macro-averaged variants. For this shared task baselines have not been reported.
Consequently, the evaluation focuses exclusively on comparing the submitted systems based on the reported metrics, allowing for a direct assessment of each approach’s relative performance.

\subsection{Evaluation Metrics}
\label{sec:metrics}
The criteria used to evaluate the submitted work reflect standard evaluation practices for assessing system performance and methodological soundness, and are as follows:
 \[
    Precision = \frac{TP}{TP + FP}
  \]

  \[
    Recall = \frac{TP}{TP + FN}
  \]

  \[
    F1 = \frac{2 \cdot Precision \cdot Recall}{Precision + Recall}
  \]
  \\
  For any metric $M \in \{\text{Precision}, \text{Recall}, \text{F1}\}$, the macro-average is defined as:
\[
\text{Macro-avg}(M) = \frac{M_{\text{class }0} + M_{\text{class }1}}{2}.
\]

\subsection{Team Results}
The outcomes from our application of the proposed method we’ve described will now be discussed in
detail. The performance of our system is assessed with the metrics shown in Section~\ref{sec:metrics}. As shown in
the following Table~\ref{tab:classification_results}, our system achieved the reported results.
These results provide insight into the effectiveness of the proposed approach across different evaluation metrics.

\begin{table}[h!]
\centering
\caption{Classification performance per language}
\label{tab:classification_results}
\begin{tabular}{l|ccc|ccc|ccc}
\hline
 & \multicolumn{3}{c|}{Class 0} & \multicolumn{3}{c|}{Class 1} & \multicolumn{3}{c}{Macro-avg} \\
Language 
& Precision & Recall & F1-score 
& Precision & Recall & F1-score 
& Precision & Recall & F1-score \\
\hline
English & 0.9355 &  0.9281 &  0.9318 & 0.2968 & 0.3220 & 0.3089 & 0.6162 & 0.6250 & 0.6203   \\
Italian & 0.9450 & 0.9692 & 0.9570 & 0.8548 & 0.7625 & 0.8060 & 0.8999 & 0.8659 & 0.8815  \\
Spanish &  0.9707 & 0.7344 & 0.8361 & 0.3684 & 0.875 & 0.5185 & 0.6695 & 0.8047 & 0.6773 \\
\hline
\end{tabular}
\end{table}

\begin{figure*}[h]
    \centering
    
    \includegraphics[width=0.7\textwidth]{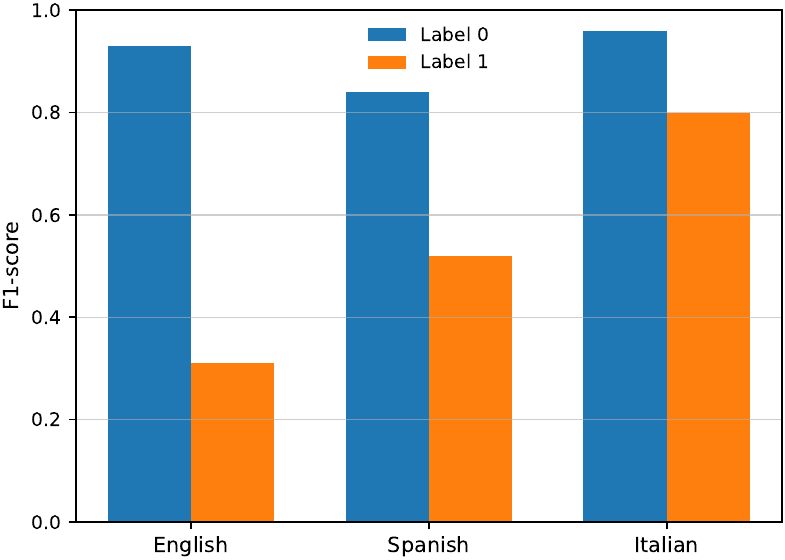}
    \caption{Comparison of F1-scores for binary classification labels (Label 0 and Label 1) across English, Spanish, and Italian.}
    \label{fig:bar}
\end{figure*}

As the main evaluation metric for system evaluation is macro F1-score, the corresponding bar plot highlighting the percentages is shown in Figure~\ref{fig:bar}, which facilitates an easier comparison of results across languages and class labels.

\section{Discussion}

The findings presented herein indicate how efficiently the proposed technique operates in the context of datasets with significant amounts of data skewness towards one particular label or another based upon the assumption that some labels will appear more frequently than others. If implemented into operational environments today, such implementations would require minimal amounts of storage space for the trained model as well as only small amounts of computation time for each individual example within each batch of processed records. Thus, our results demonstrate that lightweight algorithms can effectively learn relationships between classes/labels based solely on a very small number of records within a given training set versus other possible approaches. In conclusion, the findings presented herein can be used as a starting point for additional research involving the development and implementation of lightweight models and algorithms that incorporate this approach. Although this methodology performed well on one labeled sample and had lower accuracy on the other labeled sample, this difference in performance is an indication of the significant class imbalance in the existing dataset. In fact, the results from this methodology were consistent with what is expected from an imbalanced learning environment, which does not indicate that the model was unstable. Also, the lower performance on a second labeled sample does not mean that the model is lacking capability but is indicative of the imbalanced data distribution and demonstrates that when looking at evaluation metrics, one must take into account the nature of the data.
\section{Conclusions}

Our team believes that utilizing a minimal alternative provides a new path for solving the problem of distinguishing between hate speech and reclaimed speech/language. This method provides an ideal balance between effectiveness and efficiency when dealing with limited training resources and computational capabilities. While our proposed approach performs well with limited resources, it has produced strong results across the evaluation criteria, although performance varies across languages; in particular, results for English were lower compared to other languages. Therefore, the applicability of this methodology across multiple languages demonstrates the potential strength and generalizability of our method in a multilingual environment. Moving forward, it will be valuable to utilize larger language models or generative models, as well as additional advanced methods and more complex learning approaches, to address remaining classification difficulties and ultimately enhance the overall accuracy of the system.


\section*{Declaration on Generative AI}
  The application of generative artificial intelligence (AI) tools occurred only in a limited capacity to support the research undertaken in this study. For example, generative AI provided assistance with translating and interpreting meaning from the tweet dataset containing Italian language tweets as the authors of the study do not have fluency in Italian, being native speakers of English. In terms of facilitating improvement and clarity in the transcription of notes and written work generated by the authors, generative AI was also used in this process. Although generative AI was not applied towards generating, annotating, modelling or analysing any of the data collected or interpreting results from this study, all scientific/ statistical decisions and interpretations made throughout the study were conducted exclusively by the authors. After using these tools/services, the authors reviewed and edited the content as needed and takes full responsibility for the publication’s content.  

\bibliography{refs}


\end{document}